\pdfoutput=1

\documentclass[11pt]{article}

\usepackage[final]{acl}

\usepackage{times}
\usepackage{latexsym}

\usepackage[T1]{fontenc}

\usepackage[utf8]{inputenc}

\usepackage{microtype}

\usepackage{inconsolata}

\usepackage{graphicx}
\usepackage{siunitx} %

\usepackage{listings}
\usepackage{tikz}
\usepackage{pgfplots}
\usepackage{booktabs}
\usepackage{stfloats}
\usepackage{multirow}
\usepackage[most]{tcolorbox}
\usepackage{amssymb}
\usepackage{amsmath}
\usepackage{amsthm}
\usepackage{enumitem}

\newtheorem*{theorem}{Theorem}

\pgfplotsset{compat=1.18}

\usepackage{color}
\definecolor{keywordcolor}{rgb}{0.7, 0.1, 0.1}   %
\definecolor{tacticcolor}{rgb}{0.0, 0.1, 0.6}    %
\definecolor{commentcolor}{rgb}{0.4, 0.4, 0.4}   %
\definecolor{symbolcolor}{rgb}{0.0, 0.1, 0.6}    %
\definecolor{sortcolor}{rgb}{0.1, 0.5, 0.1}      %
\definecolor{attributecolor}{rgb}{0.7, 0.1, 0.1} %

\newcommand{\sparagraph}[1]{\noindent\textbf{#1.}\ }

\title{RLMEval: Evaluating Research-Level Neural Theorem Proving}

\author{Auguste Poiroux \quad Antoine Bosselut \quad Viktor Kunčak \\
School of Computer and Communication Sciences \\
EPFL, Switzerland \\
\texttt{\{auguste.poiroux, antoine.bosselut, viktor.kuncak\}@epfl.ch}}

\begin{document}
\maketitle

\begin{abstract}
    Despite impressive results on curated benchmarks, the practical impact of large language models (LLMs) on research-level neural theorem proving and proof autoformalization is still limited. We introduce \href{https://github.com/augustepoiroux/RLMEval}{\textbf{RLMEval}}, an evaluation suite for these tasks, focusing on research-level mathematics from real-world Lean formalization projects. RLMEval targets the evaluation of neural theorem proving and proof autoformalization on challenging research-level theorems by leveraging real Lean Blueprint formalization projects.
    Our evaluation of state-of-the-art models on RLMEval, comprising 613 theorems from 6 Lean projects, reveals a significant gap: progress on existing benchmarks does not readily translate to these more realistic settings, with the best model achieving only a \SI{10.3}{\percent} pass rate. RLMEval provides a new, challenging benchmark designed to guide and accelerate progress in automated reasoning for formal mathematics.
\end{abstract}

\section{Introduction}
Automatically translating mathematical content from natural language into a formal language suitable for proof assistants \citep{benzmuller_promising_2020, wang_exploration_2020}, is crucial for bridging human mathematical reasoning with machine-verifiable proofs.
Neural theorem proving (NTP) and proof autoformalization using large language models (LLMs) have demonstrated remarkable progress \citep{polu_generative_2020, jiang_draft_2023} in these tasks. This success is predominantly measured on curated benchmarks such as MiniF2F \citep{zheng_minif2f_2022} or ProofNet \citep{azerbayev_proofnet_2023}. While valuable, these benchmarks suffer from issues, such as saturation (e.g., MiniF2F reaching \SI{88.9}{\percent} success \citep{ren_deepseek-prover-v2_2025}), formalization inaccuracies (e.g., ${\sim}\SI{30}{\percent}$ in ProofNet \citep{poiroux_improving_2024}), or a narrow focus on competition-style problems. They do not fully capture the complexities of real-world, research-level mathematics. Consequently, model performance on these benchmarks may not reliably predict their practical utility in assisting with ongoing, complex formalization projects.

To address this gap, this work introduces \textbf{RLMEval}\footnote{Apache 2.0 license, similar to the projects it relies on}, a benchmark for evaluating neural theorem proving and proof autoformalization on research-level mathematics within contemporary Lean 4 projects. RLMEval distinguishes itself by focusing on \emph{blueprint theorems}, significant, high-level results from real-world Lean projects. These theorems embody core conceptual advances, unlike the more numerous auxiliary lemmas which often involve smaller, routine deductions and typically constitute over 75\% of theorems in Lean projects (see \autoref{tab:project_proof_length_stats}). By concentrating on these challenging blueprint theorems, RLMEval provides a more realistic and demanding testbed for LLMs.\footnote{Examples from RLMEval are in \autoref{sec:appendix}.} This targeted evaluation aims to steer LLM development towards capabilities that can meaningfully contribute to the advancement of formal mathematics. Our main contributions are the following:
\begin{enumerate}[leftmargin=*]\itemsep=0pt
    \item RLMEval: A novel evaluation benchmark for research-level neural theorem proving and proof autoformalization. RLMEval is applicable to a wide range of Lean blueprint projects and, to our knowledge, is the first benchmark specifically designed for these advanced tasks at the research level within the Lean ecosystem. Extensible and versioned annually, RLMEval will continuously test model capabilities and limit data contamination.
    \item An evaluation of state-of-the-art LLMs using RLMEval. This evaluation reveals a significant disparity in performance compared to established benchmarks, thereby pinpointing critical areas for future research and development.
\end{enumerate}

\section{Related Work}
\label{sec:related_work}

Research in neural theorem proving (NTP) has seen significant advancements, from early work like GPT-f \citep{polu_generative_2020} to recent LLM-based techniques \citep[e.g.,][]{xin_deepseek-prover_2024, ren_deepseek-prover-v2_2025, wang_kimina-prover_2025}. Some methods explore using intermediate, possibly less-rigorous reasoning to guide formal proof generation, while others, like LeanAgent \citep{kumarappan_leanagent_2025}, train on multiple Lean repositories in a curriculum to accumulate knowledge.
Proof autoformalization has also progressed, with early breakthroughs in Isabelle where models like Codex translated informal math problems into formal specifications \citep{wu_autoformalization_2022}. The Draft, Sketch, and Prove (DSP) approach \citep{jiang_draft_2023} further improved this by using informal proofs to generate formal proof sketches, which are then completed by an automated theorem prover.

\textsc{MiniF2F} \citep{zheng_minif2f_2022} offers Olympiad-level problems across multiple proof assistants. \textsc{ProofNet} \citep{azerbayev_proofnet_2023} provides Lean theorems with informal statements/proofs, targeting both autoformalization and theorem proving. \textsc{PutnamBench} \citep{tsoukalas_putnambench_2024} focuses on challenging Putnam competition problems. These benchmarks are focused on competition-style problems or undergraduate-level mathematics. \textsc{miniCTX} \citep{hu_minictx_2024} evaluates provers on real Lean projects but is tied to a specific Lean version and includes all theorems, many of which are auxiliary technical lemmas.

RLMEval differentiates itself by: (1) focusing on \emph{research-level mathematics} drawn from real-world Lean blueprint projects \citep{massot_patrickmassotleanblueprint_2025}; (2) specifically targeting \emph{blueprint theorems}, which represent significant conceptual steps, unlike benchmarks that include numerous simpler lemmas (see \autoref{tab:project_proof_length_stats}); (3) introducing \emph{proof autoformalization} (informal proof to formal proof) alongside neural theorem proving as core tasks; and (4) ensuring broad compatibility with Lean versions, allowing evaluation on a wider array of ongoing projects.

\begin{table*}[h!tb]
    \small
    \centering
    \caption{Lean Blueprint projects used to build RLMEval. We obtained agreement from the primary authors of these projects to evaluate our models on them. Difficulty is a subjective assessment based on our experience with the projects and the models' performance.}
    \begin{tabular}{lcccrr}
        \toprule
        \textbf{Project}                                                                                                   & \textbf{Domain}                   & \textbf{Difficulty} & \textbf{\#Thms} & \textbf{Lean} & \textbf{First Commit} \\ \midrule
        \href{https://github.com/fpvandoorn/carleson/tree/ec175b9008144d009269ce427b9ad43dbd70d0a5}{Carleson}              & Analysis                          & hard                & 110             & v4.14.0-rc2   & 20 Oct 2023           \\
        \href{https://github.com/imperialcollegelondon/FLT/tree/fed5e57b05e232f3bfe24d24098111e9dcd7bcd1}{FLT}             & Number Theory                     & medium              & 52              & v4.14.0-rc2   & 19 Nov 2023           \\
        \href{https://github.com/teorth/pfr/tree/f6bdcac2365623d3667d3ff8fd8ddb7f95ce2313}{PFR}                            & Combinatorics                     & hard                & 144             & v4.14.0-rc3   & 13 Nov 2023           \\
        \href{https://github.com/alexkontorovich/PrimeNumberTheoremAnd/tree/6101a4b1f0cd4096b0c41cc90c7ba89f7593ef77}{PNT} & Analytic Number Theory            & medium              & 99              & v4.14.0-rc2   & 9 Jan 2024            \\
        \href{https://github.com/pitmonticone/FLT3/tree/a199fa0467f86504a9d2f6164b0456608e586821}{FLT3}                    & Number Theory                     & easy                & 84              & v4.7.0-rc2    & 22 Mar 2024           \\
        \href{https://github.com/remydegenne/testing-lower-bounds/tree/0f09ff100a06a5e4542181514bfff74213ae126b}{TLB}      & Information \& Probability Theory & medium              & 124             & v4.13.0-rc3   & 22 Feb 2024           \\ \bottomrule
    \end{tabular}
    \label{tab:real_project_details}
\end{table*}

\section{Methodology}
\label{sec:methodology}

RLMEval provides a comprehensive suite for evaluating neural theorem proving and proof autoformalization on research-level Lean mathematics. We release it along with a dedicated Python interface, \href{https://github.com/augustepoiroux/LeanInteract}{LeanInteract}, for robust communication with the Lean proof assistant \citep{10.1007/978-3-030-79876-5_37}, offering programmatic control over Lean through its REPL interface \cite{noauthor_leanprover-communityrepl_2025}. A key feature crucial for RLMEval is its multi-version support, achieved through manual backporting of the latest REPL features and bug fixes to all 41 Lean versions between v4.7.0-rc1 and v4.19.0 (a significant undertaking ensuring broad applicability). This approach prioritizes broad compatibility and ease of use for benchmarking across the evolving Lean ecosystem, setting it apart from existing interaction tools like LeanDojo \citep{yang_leandojo_2023} or Pantograph \citep{aniva_pantograph_2024}, which are tied to specific Lean versions or require compute-intensive project tracing. This property is a key cornerstone of RLMEval for its extensibility, laying the foundations for long-term maintenance and future updates. Because our benchmark is based on real-world projects, data contamination concerns apply. In order to continuously mitigate this risks, we plan to release new versions of RLMEval with more recent Lean projects to continuously reduce data contamination risks when evaluating current and future models.

\paragraph{Benchmark Design.}

\begin{table*}[t]
    \small
    \centering
    \caption{Proof length statistics (in number of lines, comments are trimmed) for theorems within project blueprints (main theorems) versus auxiliary lemmas across the RLMEval projects. The '\% Auxiliary lemmas' column indicates the proportion of all theorems that are auxiliary lemmas.}
    \begin{tabular}{@{}lccc@{}}
        \toprule
        \textbf{Project} & \textbf{\% Auxiliary lemmas} & \textbf{Main theorems Proof Length} & \textbf{Auxiliary lemmas Proof Length} \\
        \midrule
        PFR              & \SI{75.3}{\percent}          & 23.2                                & 9.0                                    \\
        FLT              & \SI{72.2}{\percent}          & 12.8                                & 4.0                                    \\
        FLT3             & \SI{15.9}{\percent}          & 8.8                                 & 4.9                                    \\
        Carleson         & \SI{85.9}{\percent}          & 27.0                                & 7.8                                    \\
        PNT              & \SI{78.3}{\percent}          & 16.7                                & 8.3                                    \\
        TLB              & \SI{83.0}{\percent}          & 11.2                                & 5.7                                    \\
        Avg              & \SI{68.3}{\percent}          & 16.6                                & 6.6                                    \\
        \bottomrule
    \end{tabular}
    \label{tab:project_proof_length_stats}
\end{table*}

Our methodology for curating RLMEval is inspired by \textsc{RLM25} \citep{poiroux_improving_2024}, a research-level benchmark for autoformalization of theorem statements.
RLMEval leverages existing Lean blueprint projects \citep{massot_patrickmassotleanblueprint_2025}, which curate high-quality alignments between natural language mathematics and their formal counterparts. These projects, authored by domain experts, ensure the precision and real-world relevance of the benchmark data.

A core design principle of RLMEval is its focus on \emph{blueprint theorems}, i.e. formal theorems that are linked in the informal blueprint of the projects. The blueprint theorems represent the main, high-level steps of mathematical development, akin to theorems found in research papers. This contrasts with previous works \citep{kumarappan_leanagent_2025, hu_minictx_2024} that include a large proportion of simpler, auxiliary lemmas. \autoref{tab:project_proof_length_stats} illustrates the difference in proof lengths between the main theorems and auxiliary lemmas in RLMEval: 16.6 vs 6.6 lines on average. Auxiliary lemmas regularly represent more than \SI{75}{\percent} of the total theorems in a Lean project. By focusing on blueprint theorems, RLMEval targets more complex, research-level reasoning tasks, providing a more challenging and realistic benchmark.
RLMEval comprises 613 theorems from the 6 Lean projects detailed in \autoref{tab:real_project_details}.

\noindent RLMEval supports the following tasks:

\begin{itemize}[leftmargin=*]
    \item \textbf{Neural Theorem Proving (NTP)}: Given a formal Lean statement, generate a complete and verifiable Lean proof.
    \item \textbf{Proof Autoformalization}: Given an informal (natural language) proof and its corresponding formal statement, generate a complete, verifiable Lean proof.
\end{itemize}

\noindent To assess models under varying conditions, RLMEval uses two evaluation modes:

\begin{itemize}[leftmargin=*]
    \item \textbf{Easy mode}: Models access all definitions and lemmas from the source project, including non-blueprint auxiliary lemmas, i.e. technical lemmas that are not included in the informal blueprint.
    \item \textbf{Normal mode}: Models access only blueprint theorems from the source project; they do not have access to these non-blueprint auxiliary lemmas and thus may need to prove equivalent intermediate results themselves. This simulates a more realistic task, closer to what mathematicians formalizing research results are faced with.
\end{itemize}

\section{Experimental Setup and Results}
\label{sec:results}

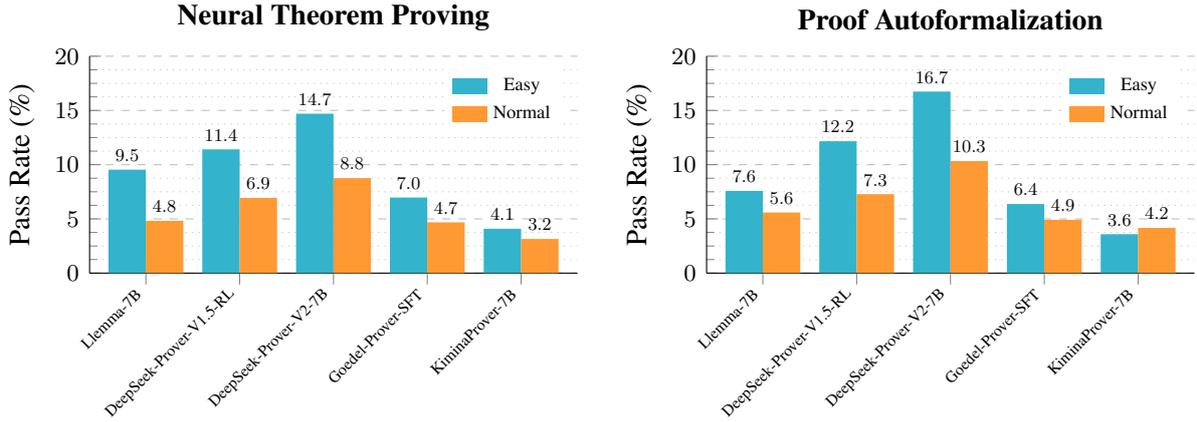
\begin{figure*}[ht]
    \centering
    \definecolor{easycolor}{rgb}{0.20,0.70,0.80}   %
\definecolor{normalcolor}{rgb}{1.00,0.60,0.20} %

\begin{tikzpicture}
  \begin{axis}[
      name=ax1,
      title=\textbf{Neural Theorem Proving},
      bar width=14pt,
      ybar=0pt,
      width=0.5\linewidth,
      height=0.18\textheight,
      enlarge x limits=0.15,
      symbolic x coords={Llemma-7B, DeepSeek-Prover-V1.5-RL, DeepSeek-Prover-V2-7B, Goedel-Prover-SFT, KiminaProver-7B},
      xtick=data,
      x tick label style={font=\tiny, rotate=45, anchor=east},
      ymin=0, ymax=20,
      ytick={0,5,10,15,20},
      yticklabel style={font=\footnotesize},
      ylabel={Pass Rate (\%)},
      axis lines*=left,
      ymajorgrids,
      grid style={dashed},
      yminorgrids=true,
      minor y tick num=3,      %
      minor grid style={dotted},
      nodes near coords,
      every node near coord/.append style={
          font=\small,            %
          scale=0.75,             %
          /pgf/number format/fixed,
          /pgf/number format/zerofill,
          /pgf/number format/precision=1,
          anchor=south,
          yshift=0.5pt,
          text=black
        },
      area legend,
      legend style={
          font=\scriptsize,
          at={(0.85,0.95)},
          anchor=north,
          draw=none,
        },
      legend image post style={xscale=0.75}
    ]

    \addplot[fill=easycolor, draw opacity=0] coordinates {
        (Llemma-7B, 9.54)
        (DeepSeek-Prover-V1.5-RL, 11.41)
        (DeepSeek-Prover-V2-7B, 14.7)
        (Goedel-Prover-SFT, 6.97)
        (KiminaProver-7B, 4.09)
      };
    \addplot[fill=normalcolor, draw opacity=0] coordinates {
        (Llemma-7B, 4.82)
        (DeepSeek-Prover-V1.5-RL, 6.94)
        (DeepSeek-Prover-V2-7B, 8.76)
        (Goedel-Prover-SFT, 4.68)
        (KiminaProver-7B, 3.15)
      };

    \legend{Easy, Normal}
  \end{axis}

  \begin{axis}[
      at={(ax1.south east)},
      name=ax2,
      title=\textbf{Proof Autoformalization},
      bar width=14pt,
      ybar=0pt,
      width=0.5\linewidth,
      height=0.18\textheight,
      xshift=1.7cm,
      enlarge x limits=0.15,
      symbolic x coords={Llemma-7B, DeepSeek-Prover-V1.5-RL, DeepSeek-Prover-V2-7B, Goedel-Prover-SFT, KiminaProver-7B},
      xtick=data,
      x tick label style={font=\tiny, rotate=45, anchor=east},
      ymin=0, ymax=20,
      ytick={0,5,10,15,20},
      yticklabel style={font=\footnotesize},
      ylabel={Pass Rate (\%)},
      axis lines*=left,
      ymajorgrids,
      grid style={dashed},
      yminorgrids=true,
      minor y tick num=3,      %
      minor grid style={dotted},
      nodes near coords,
      every node near coord/.append style={
          font=\small,
          scale=0.75,             %
          /pgf/number format/fixed,
          /pgf/number format/zerofill,
          /pgf/number format/precision=1,
          anchor=south,
          yshift=0.5pt,
          text=black
        },
      area legend,
      legend style={
          font=\scriptsize,
          at={(0.85,0.95)},
          anchor=north,
          draw=none,
        },
      legend image post style={xscale=0.75}
    ]

    \addplot[fill=easycolor, draw opacity=0] coordinates {
        (Llemma-7B, 7.58)
        (DeepSeek-Prover-V1.5-RL, 12.16)
        (DeepSeek-Prover-V2-7B, 16.74)
        (Goedel-Prover-SFT, 6.37)
        (KiminaProver-7B, 3.59)
      };

    \addplot[fill=normalcolor, draw opacity=0] coordinates {
        (Llemma-7B, 5.59)
        (DeepSeek-Prover-V1.5-RL, 7.28)
        (DeepSeek-Prover-V2-7B, 10.33)
        (Goedel-Prover-SFT, 4.91)
        (KiminaProver-7B, 4.17)
      };

    \legend{Easy, Normal}
  \end{axis}

\end{tikzpicture}
    \caption{Pass rate on RLMEval using pass@128 for neural theorem proving (left) and proof autoformalization (right), in Easy and Normal modes.}
    \label{fig:main_results}
\end{figure*}

We evaluate several LLMs on the RLMEval tasks. Our primary metric is \texttt{pass@k}, which measures the percentage of problems solved with at least one successful proof among $k$ generated attempts. For our main results, we use $k=128$ samples per problem, a budget significantly larger than the 8 samples used in miniCTX \citep{hu_minictx_2024}, allowing for a more comprehensive assessment. For each theorem, models receive the full in-file context up to the point where the proof is to be generated, including all preceding definitions, lemmas, and imports within that specific file. In normal mode, lemmas not present in the informal blueprint are excluded from the context, while in easy mode, all lemmas from the project are available.

Our baseline model is \textbf{Llemma 7B} \citep{azerbayev_llemma_2023}, a model pretrained on mathematics, but not for proof-search specifically. We also evaluate leading models specifically tuned for Lean theorem proving: \textbf{DeepSeek-Prover-V1.5-RL} \citep{xin_deepseek-prover-v15_2024}, \textbf{DeepSeek-Prover-V2-7B} \citep{ren_deepseek-prover-v2_2025}, \textbf{Goedel-Prover-SFT} \citep{lin_goedel-prover_2025}, and \textbf{KiminaProver-Preview-7B} \citep{wang_kimina-prover_2025}. Our evaluation on RLMEval examines their ability to generalize to research-level mathematics.

\autoref{fig:main_results} presents the main pass@128 rates for both neural theorem proving and proof autoformalization tasks, comparing ``easy'' and ``normal'' modes. The overall performance on RLMEval is markedly lower than on benchmarks like MiniF2F. For instance, DeepSeek-Prover-V2-7B, the best-performing model, achieves only \SI{10.3}{\percent} pass@128 on proof autoformalization (normal mode), in stark contrast to its reported \SI{75}{\percent}+ on MiniF2F with a smaller pass@32 budget. This disparity underscores that \textbf{current models struggle significantly with the complexity of research-level mathematics in real-world projects}.

Models consistently perform better in the ``easy'' mode (with access to project-specific auxiliary lemmas). DeepSeek-Prover-V2-7B improves from \SI{8.8}{\percent} (normal) to \SI{14.7}{\percent} (easy) in neural theorem proving (NTP), and from \SI{10.3}{\percent} to \SI{16.7}{\percent} in proof autoformalization. This performance gap highlights the challenges models face when working without direct access to project-specific lemma support.

\textbf{Providing an informal proof offers a modest benefit} (proof autoformalization versus neural theorem proving (NTP)). For DeepSeek-Prover-V2-7B (normal mode), this translates to an improvement of $\sim$\SI{1.5}{} percentage points (from \SI{8.8}{\percent} to \SI{10.3}{\percent}). This suggests that current models can leverage informal proofs to some extent, but the benefit remains limited for complex, research-level problems.

\textbf{Performance varies substantially across RLMEval projects} (see \autoref{tab:full_results} for details). For example, FLT3 yields the highest success rates (up to \SI{32.1}{\percent} for DeepSeek-Prover-V2-7B), while Carleson presents a greater challenge (max \SI{2.73}{\percent}). This variation indicates that mathematical domain and formalization style significantly impact model effectiveness.

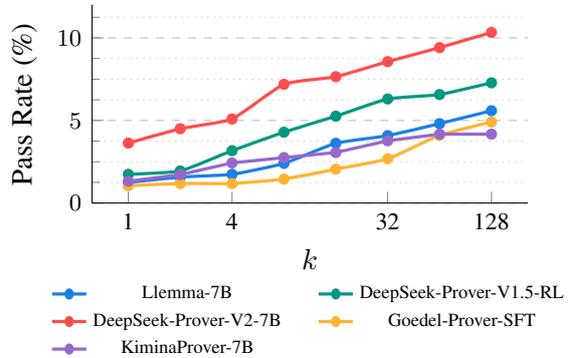
\begin{figure}[t]
    \centering
    \definecolor{colorLlemma}{rgb}{0.1,0.5,0.9} %
\definecolor{colorDSv15}{rgb}{0,0.6,0.5} %
\definecolor{colorDSv2}{rgb}{1,0.3,0.3} %
\definecolor{colorGoedel}{rgb}{1,0.7,0.2} %
\definecolor{colorKimina}{rgb}{0.6,0.4,0.8} %

\begin{tikzpicture}
    \begin{axis}[
            width=0.95\columnwidth,
            height=0.17\textheight,
            xlabel={$k$},
            ylabel={Pass Rate (\%)},
            xtick={1, 4, 32, 128},
            xticklabels={1, 4, 32, 128},
            x tick label style={font=\small},
            ytick={0, 5, 10, 15},
            yticklabel style={font=\small},
            xlabel near ticks,
            ylabel near ticks,
            ymin=0, ymax=12,
            axis lines*=left,
            ymajorgrids,
            grid style={dashed},
            yminorgrids=true,
            minor y tick num=3,
            minor grid style={dotted},
            xmode=log,
            legend style={
                    at={(0.5,-0.36)},
                    anchor=north,
                    legend columns=2,
                    font=\scriptsize,
                    draw=none,
                    /tikz/every even column/.append style={column sep=0.4cm}
                },
            legend image post style={xscale=0.75},
            trim axis left,
        ]
        \addplot[draw=colorLlemma, mark=*, mark size=1.5pt, thick, line width=1.2pt, mark options={fill=colorLlemma, draw=colorLlemma}, smooth, tension=0.3] coordinates {
                (1, 1.25) (2, 1.57) (4, 1.74) (8, 2.40) (16, 3.62) (32, 4.08) (64, 4.81) (128, 5.59)
            };

        \addplot[draw=colorDSv15, mark=*, mark size=1.5pt, thick, line width=1.2pt, mark options={fill=colorDSv15, draw=colorDSv15}, smooth, tension=0.3] coordinates {
                (1, 1.74) (2, 1.94) (4, 3.17) (8, 4.29) (16, 5.26) (32, 6.30) (64, 6.57) (128, 7.28)
            };

        \addplot[draw=colorDSv2, mark=*, mark size=1.5pt, thick, line width=1.2pt, mark options={fill=colorDSv2, draw=colorDSv2}, smooth, tension=0.3] coordinates {
                (1, 3.63) (2, 4.50) (4, 5.09) (8, 7.19) (16, 7.65) (32, 8.56) (64, 9.41) (128, 10.33)
            };

        \addplot[draw=colorGoedel, mark=*, mark size=1.5pt, thick, line width=1.2pt, mark options={fill=colorGoedel, draw=colorGoedel}, smooth, tension=0.3] coordinates {
                (1, 1.05) (2, 1.18) (4, 1.18) (8, 1.45) (16, 2.05) (32, 2.68) (64, 4.10) (128, 4.91)
            };

        \addplot[draw=colorKimina, mark=*, mark size=1.5pt, thick, line width=1.2pt, mark options={fill=colorKimina, draw=colorKimina}] coordinates {
                (1, 1.32) (2, 1.72) (4, 2.43) (8, 2.75) (16, 3.06) (32, 3.77) (64, 4.17) (128, 4.17)
            };

        \legend{Llemma-7B, DeepSeek-Prover-V1.5-RL, DeepSeek-Prover-V2-7B, Goedel-Prover-SFT, KiminaProver-7B}
    \end{axis}
\end{tikzpicture}
    \caption{Scaling trends for the proof autoformalization task in normal mode on RLMEval for various models and pass@$k$ values.}
    \label{fig:scaling}
\end{figure}

\autoref{fig:scaling} illustrates how performance scales with an increasing number of samples per theorem for various models. As in traditional benchmarks, we observe that pass rate improves with additional samples. While pass rates continue to increase with more samples for most models, the gains diminish noticeably, suggesting that simply scaling up sampling may not overcome the fundamental challenges posed by research-level mathematics. This contrasts with results on benchmarks like miniF2F, where aggressive sampling (e.g., pass@8192) produced substantial improvements \cite{ren_deepseek-prover-v2_2025}, further highlighting the greater difficulty of RLMEval.

\sparagraph{Proof length}
Proof length serves as a natural proxy for theorem complexity and human-perceived difficulty. \autoref{tab:project_proof_length_stats} reveals a clear difficulty ordering based on average proof lengths: FLT3 < TLB < FLT < PNT < PFR < Carleson. This ordering strongly correlates with model performance trends shown in \autoref{tab:full_results} and \autoref{tab:full_results_ntp}, where pass@k scores consistently degrade from FLT3 to Carleson across all evaluated models. This correlation validates proof length as a meaningful indicator of difficulty for both humans and automated theorem provers.

Analysis of successful model-generated proofs reveals significant insights about current capabilities. \autoref{tab:proof_length} shows that models successful proofs are short compared to their human-written counterparts, averaging only 2.5--6.0 lines compared to the 16.6-line human average. Critically, our inspection reveals that virtually all LLM-generated proofs are longer than the corresponding human proofs, indicating that current models primarily succeed on theorems that admit concise proof strategies, a subset representing the easier problems within RLMEval.

DeepSeek-Prover-V2-7B exhibits notably different behavior, generating proofs averaging 6.0 lines compared to 2.5--2.8 lines for other models, and producing the longest individual proof at 111 lines. Manual inspection reveals this model tends to generate verbose proofs with repetitive or redundant steps, suggesting room for improvement in proof conciseness and efficiency.

\begin{table}[h!tb]
    \centering
    \begin{tabular}{@{}lrr@{}}
        \toprule
        \textbf{Model}            & \textbf{Average} & \textbf{Max} \\
        \midrule
        Llemma-7B                 & 2.8              & 20           \\
        DeepSeek-Prover-V1.5-RL   & 2.6              & 20           \\
        DeepSeek-Prover-V2-7B     & 6.0              & 111          \\
        Goedel-Prover-SFT         & 2.5              & 14           \\
        KiminaProver-7B           & 2.6              & 14           \\
        Human-Written             & 16.6             & 248          \\
        \bottomrule
    \end{tabular}
    \caption{Proof length in lines of code, comments trimmed, for successful proofs generated by models on RLMEval. The length of the official human-written proofs is reported for comparison. Data aggregated from the proof autoformalization, normal mode experiments.}
    \label{tab:proof_length}
\end{table}

\section{Conclusion}
We introduced \textbf{RLMEval}, a benchmark for neural theorem proving and proof autoformalization on research-level mathematics. Focusing on real-world mathematical formalization challenges, RLMEval establishes a  realistic standard for evaluating the practical utility of neural theorem provers in mathematical settings.
Our evaluation on 613 research-level mathematics theorems reveals a stark performance drop compared to traditional benchmarks like MiniF2F. The best model achieved only \SI{10.3}{\percent} on proof autoformalization (normal mode), highlighting that current models are still far from reliably handling the complexities of ongoing formalization efforts. This performance gap is particularly pronounced for projects like Carleson with complex mathematical structures, while being somewhat narrower for more concrete domains like algebraic number theory (FLT3).
The disparity between ``easy'' and ``normal'' modes across all models demonstrates the critical role of auxiliary lemmas in successful theorem proving. Access to these intermediate results improved performance by up to \SI{6}{}\%, suggesting that enhanced techniques for lemma discovery or generation could be valuable directions for future research.

\section*{Limitations}

\sparagraph{Potential data contamination} Llemma 7B \citep{azerbayev_llemma_2023} has been released before the first commit of the projects used in RLMEval, making data leakage impossible. However, other models have been released more recently, and data contamination is unclear given the lack of information about the content of their pre-training set. Note, however, that data contamination would likely result in overestimating the current state of the art, which we already found to be low compared to more artificial benchmarks. Furthermore, thanks to the use of the LeanInteract tool \cite{poiroux_leaninteract_2025}, RLMEval is extensible to new Lean blueprint projects. As such, we plan to release new versions of RLMEval with more recent Lean projects to continuously reduce data contamination risks when evaluating current and future models.

\sparagraph{Limited Scaling} We evaluate using pass@128 which, while being substantially higher than the pass@8 from miniCTX \citep{hu_minictx_2024}, falls short to the pass@8192 on MiniF2F or pass@1024 on ProofNet used by DeepSeek in \citet{ren_deepseek-prover-v2_2025}. Additionally, we use the 7B version of DeepSeek-Prover-V2 and not the 671B one. Our results therefore very likely underestimate the best achievable performance on RLMEval as of today. Nevertheless, the 7B model already achieves \SI{75.6}{\percent} in a low compute budget (pass@32) on MiniF2F while the 671B version achieves \SI{88.9}{\percent} on the latest sampling budget (pass@8192). These results indicate that even the 671B model would likely not achieve high scores on RLMEval. Additionally, running such a large model with pass@8192 requires a large amount of computation and time which, as of today, are a bit unrealistic for practical usage.

\sparagraph{Context} Our current evaluation provides models with the full in-file context preceding the target theorem. While common, this setup may disadvantage neural theorem proving (NTP) models trained on self-contained theorems, making them less adept at leveraging long, complex in-file contexts. Ideally, models should receive a more sophisticated context, incorporating broader project and research information. Future work should investigate optimal context retrieval strategies for these research-level tasks, moving beyond simple in-file truncation. Techniques could include retrieval-augmented generation or methods to summarize or filter premises from the entire project or its dependencies. Developing and evaluating such mechanisms is a key area for future work.

\sparagraph{Informal Proofs} One interesting aspect of our evaluation is how the availability of informal proofs affects performance. While we observed a modest improvement when providing informal proofs over just providing formal statements, the benefit is somewhat limited. In particular, we found that some informal proofs in RLMEval can be cryptic without broader project context, using shorthand references to other theorems or principles. The PFR sample in \autoref{sec:appendix} is such an example. These terse proofs illustrate the gap between idealized benchmark settings, using self-contained informal proofs, and real-world mathematics as depicted by RLMEval. Future work should explore how to better leverage informal proofs, possibly by providing additional context.

\section*{Acknowledgments}
We thank the Lean community for their support and feedback, in particular the authors of the Lean blueprint projects included in RLMEval. We also gratefully acknowledge the support of the IC school of computer and communication sciences, the Swiss National Science Foundation (No. 215390), Innosuisse (PFFS-21-29), the EPFL Center for Imaging, Sony Group Corporation, and a Meta LLM Evaluation Research Grant.

\bibliography{custom}

\appendix

\section{Appendix}
\label{sec:appendix}

\subsection{Model Configurations}
\label{sec:model_configs}

The evaluation of models for both Neural Theorem Proving (NTP) and Proof Autoformalization tasks was conducted using sampling-based decoding to compute pass@128.
Specific hyperparameters for each model are detailed in \autoref{tab:model_configurations}. All models were accessed via a local vLLM server on H100 80GB GPUs.

\begin{table*}[h!tb]
    \centering
    \caption{Hyperparameters for model evaluations. We use officially recommended hyperparameters for all models. All configurations used 128 samples per RLM25 entry.}
    \begin{tabular}{@{}lcccc@{}}
        \toprule
        \textbf{Model}          & \textbf{Temp.} & \textbf{Top P} & \textbf{Max Gen. Tokens} & \textbf{Max Total Tokens} \\
        \midrule
        Llemma 7B               & 0.7            & 0.95           & 1024                     & 4096                      \\
        DeepSeek-Prover-V1.5-RL & 1.0            & 0.95           & 1024                     & 4096                      \\
        DeepSeek-Prover-V2-7B   & 1.0            & 0.95           & 1024                     & 4096                      \\
        Goedel-Prover-SFT       & 1.0            & 0.95           & 1024                     & 4096                      \\
        KiminaProver-7B         & 0.6            & 0.95           & 4096                     & 16384                     \\
        \bottomrule
    \end{tabular}
    \label{tab:model_configurations}
\end{table*}

\subsection{Detailed results}

\begin{table*}[htbp!]
    \centering
    \caption{Detailed pass@k rates (\%) for Proof Autoformalization on RLMEval projects. Normal mode uses only blueprint lemmas, Easy mode uses all project lemmas. Projects are: PFR, FLT3 (Fermat's Last Theorem for n=3), Carleson (Carl.), FLT (Fermat's Last Theorem), TLB (testing-lower-bounds), PNT (Prime Number Theorem And).}
    \label{tab:full_results}
    \setlength{\tabcolsep}{4pt} %
    \begin{tabular}{@{}llcrrrrrrr@{}}
        \toprule
        Model                                    & Mode                    & p@k   & PFR  & FLT3  & Carl. & FLT   & TLB   & PNT   & Total \\
        \midrule
        \multirow{6}{*}{Llemma 7B}               & \multirow{3}{*}{Normal} & p@1   & 0.69 & 3.57  & 0.00  & 0.00  & 3.23  & 0.00  & 1.25  \\
                                                 &                         & p@32  & 0.69 & 9.52  & 0.91  & 3.85  & 6.45  & 3.03  & 4.08  \\
                                                 &                         & p@128 & 0.69 & 14.29 & 0.91  & 5.77  & 8.87  & 3.03  & 5.59  \\
        \cmidrule(lr){2-10}
                                                 & \multirow{3}{*}{Easy}   & p@1   & 0.00 & 2.38  & 0.00  & 1.92  & 4.03  & 1.01  & 1.56  \\
                                                 &                         & p@32  & 0.69 & 9.52  & 0.91  & 9.62  & 6.45  & 3.03  & 5.04  \\
                                                 &                         & p@128 & 1.39 & 13.10 & 0.91  & 15.38 & 9.68  & 5.05  & 7.58  \\
        \midrule
        \multirow{6}{*}{DeepSeek-Prover-V1.5-RL} & \multirow{3}{*}{Normal} & p@1   & 0.00 & 2.38  & 0.91  & 1.92  & 3.23  & 2.02  & 1.74  \\
                                                 &                         & p@32  & 1.39 & 19.05 & 0.91  & 9.62  & 4.84  & 2.02  & 6.30  \\
                                                 &                         & p@128 & 2.08 & 22.62 & 0.91  & 9.62  & 6.45  & 2.02  & 7.28  \\
        \cmidrule(lr){2-10}
                                                 & \multirow{3}{*}{Easy}   & p@1   & 0.69 & 5.95  & 0.00  & 0.00  & 1.61  & 2.02  & 1.71  \\
                                                 &                         & p@32  & 0.69 & 16.67 & 0.91  & 19.23 & 12.90 & 8.08  & 9.75  \\
                                                 &                         & p@128 & 3.47 & 23.81 & 0.91  & 21.15 & 14.52 & 9.09  & 12.16 \\
        \midrule
        \multirow{6}{*}{DeepSeek-Prover-V2-7B}   & \multirow{3}{*}{Normal} & p@1   & 0.69 & 11.90 & 0.00  & 5.77  & 2.42  & 1.01  & 3.63  \\
                                                 &                         & p@32  & 2.08 & 25.00 & 1.82  & 11.54 & 8.87  & 2.02  & 8.56  \\
                                                 &                         & p@128 & 2.08 & 32.14 & 2.73  & 11.54 & 10.48 & 3.03  & 10.33 \\
        \cmidrule(lr){2-10}
                                                 & \multirow{3}{*}{Easy}   & p@1   & 0.69 & 7.14  & 0.91  & 9.62  & 4.84  & 3.03  & 4.37  \\
                                                 &                         & p@32  & 3.47 & 27.38 & 1.82  & 23.08 & 19.35 & 10.10 & 14.20 \\
                                                 &                         & p@128 & 4.86 & 32.14 & 3.64  & 23.08 & 22.58 & 14.14 & 16.74 \\
        \midrule
        \multirow{6}{*}{Goedel-Prover-SFT}       & \multirow{3}{*}{Normal} & p@1   & 0.00 & 3.57  & 0.00  & 1.92  & 0.81  & 0.00  & 1.05  \\
                                                 &                         & p@32  & 0.69 & 8.33  & 0.00  & 3.85  & 3.23  & 0.00  & 2.68  \\
                                                 &                         & p@128 & 0.69 & 13.10 & 0.00  & 9.62  & 4.03  & 2.02  & 4.91  \\
        \cmidrule(lr){2-10}
                                                 & \multirow{3}{*}{Easy}   & p@1   & 0.00 & 1.19  & 0.00  & 0.00  & 0.81  & 0.00  & 0.33  \\
                                                 &                         & p@32  & 0.00 & 8.33  & 0.00  & 13.46 & 4.84  & 2.02  & 4.78  \\
                                                 &                         & p@128 & 0.69 & 10.71 & 0.00  & 17.31 & 6.45  & 3.03  & 6.37  \\
        \midrule
        \multirow{6}{*}{KiminaProver-7B}         & \multirow{3}{*}{Normal} & p@1   & 0.00 & 3.57  & 0.00  & 1.92  & 2.42  & 0.00  & 1.32  \\
                                                 &                         & p@32  & 0.00 & 10.71 & 0.00  & 7.69  & 3.23  & 1.01  & 3.77  \\
                                                 &                         & p@128 & 0.00 & 13.10 & 0.00  & 7.69  & 3.23  & 1.01  & 4.17  \\
        \cmidrule(lr){2-10}
                                                 & \multirow{3}{*}{Easy}   & p@1   & 0.00 & 3.57  & 0.00  & 1.92  & 0.00  & 0.00  & 0.92  \\
                                                 &                         & p@32  & 0.00 & 7.14  & 0.00  & 5.77  & 2.42  & 3.03  & 3.06  \\
                                                 &                         & p@128 & 0.00 & 9.52  & 0.00  & 5.77  & 3.23  & 3.03  & 3.59  \\
        \bottomrule
    \end{tabular}
\end{table*}

\begin{table*}[htbp!]
    \centering
    \caption{Detailed pass@k rates (\%) for Neural Theorem Proving (NTP) on RLMEval projects. Normal mode uses only blueprint lemmas, Easy mode uses all project lemmas. Projects are: PFR, FLT3 (Fermat's Last Theorem for n=3), Carleson (Carl.), FLT (Fermat's Last Theorem), TLB (testing-lower-bounds), PNT (Prime Number Theorem And).}
    \label{tab:full_results_ntp}
    \setlength{\tabcolsep}{4pt} %
    \begin{tabular}{@{}llcrrrrrrr@{}}
        \toprule
        Model                                    & Mode                    & p@k   & PFR  & FLT3  & Carl. & FLT   & TLB   & PNT   & Total \\
        \midrule
        \multirow{6}{*}{Llemma 7B}               & \multirow{3}{*}{Normal} & p@1   & 0.69 & 2.38  & 0.91  & 0.00  & 2.42  & 0.00  & 1.07  \\
                                                 &                         & p@32  & 0.69 & 8.33  & 0.91  & 5.77  & 6.45  & 2.02  & 4.03  \\
                                                 &                         & p@128 & 0.69 & 13.10 & 0.91  & 5.77  & 6.45  & 2.02  & 4.82  \\
        \cmidrule(lr){2-10}
                                                 & \multirow{3}{*}{Easy}   & p@1   & 0.00 & 2.38  & 0.91  & 1.92  & 1.61  & 2.02  & 1.47  \\
                                                 &                         & p@32  & 1.39 & 8.33  & 0.91  & 13.46 & 12.90 & 5.05  & 7.01  \\
                                                 &                         & p@128 & 2.08 & 11.90 & 0.91  & 21.15 & 16.13 & 5.05  & 9.54  \\
        \midrule
        \multirow{6}{*}{DeepSeek-Prover-V1.5-RL} & \multirow{3}{*}{Normal} & p@1   & 0.69 & 2.38  & 0.91  & 1.92  & 2.42  & 2.02  & 1.72  \\
                                                 &                         & p@32  & 1.39 & 14.29 & 0.91  & 5.77  & 6.45  & 3.03  & 5.31  \\
                                                 &                         & p@128 & 1.39 & 20.24 & 0.91  & 9.62  & 6.45  & 3.03  & 6.94  \\
        \cmidrule(lr){2-10}
                                                 & \multirow{3}{*}{Easy}   & p@1   & 0.69 & 2.38  & 0.00  & 7.69  & 3.23  & 2.02  & 2.67  \\
                                                 &                         & p@32  & 2.08 & 11.90 & 0.91  & 15.38 & 12.90 & 7.07  & 8.38  \\
                                                 &                         & p@128 & 2.78 & 20.24 & 0.91  & 21.15 & 15.32 & 8.08  & 11.41 \\
        \midrule
        \multirow{6}{*}{DeepSeek-Prover-V2-7B}   & \multirow{3}{*}{Normal} & p@1   & 0.69 & 11.90 & 0.91  & 0.00  & 4.03  & 1.01  & 3.09  \\
                                                 &                         & p@32  & 0.69 & 22.62 & 1.82  & 9.62  & 7.26  & 1.01  & 7.17  \\
                                                 &                         & p@128 & 0.69 & 25.00 & 2.73  & 9.62  & 10.48 & 4.04  & 8.76  \\
        \cmidrule(lr){2-10}
                                                 & \multirow{3}{*}{Easy}   & p@1   & 0.00 & 9.52  & 0.00  & 11.54 & 6.45  & 3.03  & 5.09  \\
                                                 &                         & p@32  & 2.78 & 19.05 & 1.82  & 21.15 & 17.74 & 9.09  & 11.94 \\
                                                 &                         & p@128 & 4.86 & 22.62 & 1.82  & 25.00 & 21.77 & 12.12 & 14.70 \\
        \midrule
        \multirow{6}{*}{Goedel-Prover-SFT}       & \multirow{3}{*}{Normal} & p@1   & 0.00 & 1.19  & 0.00  & 0.00  & 0.00  & 0.00  & 0.20  \\
                                                 &                         & p@32  & 0.69 & 7.14  & 0.00  & 7.69  & 4.84  & 0.00  & 3.39  \\
                                                 &                         & p@128 & 0.69 & 9.52  & 0.00  & 9.62  & 7.26  & 1.01  & 4.68  \\
        \cmidrule(lr){2-10}                      & \multirow{3}{*}{Easy}   & p@1   & 0.00 & 4.76  & 0.00  & 0.00  & 3.23  & 0.00  & 1.33  \\
                                                 &                         & p@32  & 0.69 & 7.14  & 0.91  & 15.38 & 6.45  & 2.02  & 5.43  \\
                                                 &                         & p@128 & 0.69 & 11.90 & 0.91  & 15.38 & 8.87  & 4.04  & 6.97  \\
        \midrule
        \multirow{6}{*}{KiminaProver-7B}         & \multirow{3}{*}{Normal} & p@1   & 0.00 & 3.57  & 0.00  & 5.77  & 0.81  & 0.00  & 1.69  \\
                                                 &                         & p@32  & 0.00 & 4.76  & 0.00  & 7.69  & 2.42  & 2.02  & 2.82  \\
                                                 &                         & p@128 & 0.00 & 5.95  & 0.00  & 7.69  & 3.23  & 2.02  & 3.15  \\
        \cmidrule(lr){2-10}
                                                 & \multirow{3}{*}{Easy}   & p@1   & 0.00 & 3.57  & 0.00  & 3.85  & 0.81  & 1.01  & 1.54  \\
                                                 &                         & p@32  & 0.00 & 4.76  & 0.00  & 11.54 & 3.23  & 3.03  & 3.76  \\
                                                 &                         & p@128 & 0.00 & 5.95  & 0.00  & 11.54 & 4.03  & 3.03  & 4.09  \\
        \bottomrule
    \end{tabular}
\end{table*}

\begin{figure*}[ht]
    \centering
    \input{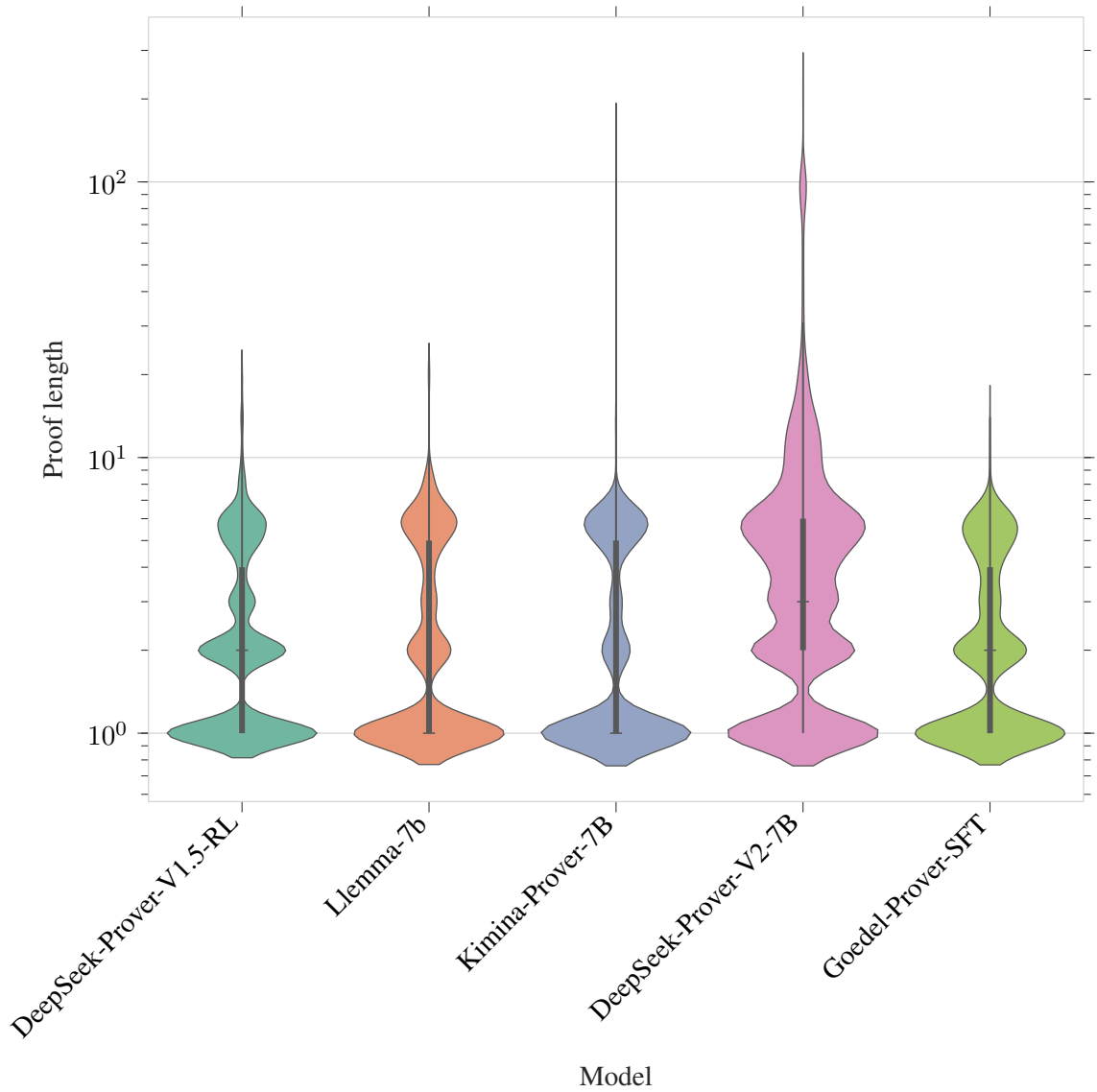}
    \caption{Proof length distribution of the different evaluated models on RLMEval. Proof length is measured in lines of code, comments trimmed. Data aggregated from all experiments: NTP and proof autoformalization, normal and easy modes.}
    \label{fig:proof_length_dist}
\end{figure*}

\clearpage
\onecolumn
\subsection{Samples from the RLMEval benchmark}

\begin{tcolorbox}[title=FLT3 sample - One of the simplest entry from RLMEval,width=\textwidth]

    \textbf{Name:} \verb|lmm:lambda_not_dvd_Y|

    \textbf{File:} \verb|FLT3/FLT3.lean|

    \begin{theorem}
        Given `S : Solution`, we have that `$\lambda$` does not divide `S.Y`.
    \end{theorem}

    \textbf{Formal statement:}

    \begin{lstlisting}
lemma lambda_not_dvd_Y : ¬ λ | S.Y
\end{lstlisting}

    \textbf{Informal proof:}

    By contradiction we assume that $\lambda \mid Y$, then, by the properties of divisibility, $\lambda \mid u_2 Y^3$, which implies, by \verb|def:Solution_u1_u2_u3_u4_u5_X_Y_Z|,
    that $\lambda \mid y$. However, this contradicts \verb|lmm:lambda_not_dvd_y| forcing us to conclude that $\lambda \nmid Y$.

    \textbf{Formal proof:}

    \begin{lstlisting}
intro h
have hyp := dvd_mul_of_dvd_right h (S.u₂ * S.Y^2)
rw [show ↑(u₂ S) * Y S ^ 2 * Y S = ↑S.u₂ * S.Y^3 by ring] at hyp
rw [← u₂_Y_spec] at hyp
apply lambda_not_dvd_y S
simp [hyp]
\end{lstlisting}

\end{tcolorbox}

\begin{tcolorbox}[title=Carleson sample - Typical difficult entry of RLMEval, breakable,width=\textwidth]

    \textbf{Name:} \verb|tile-sum-operator|

    \textbf{File:} \verb|Carleson/FinitaryCarleson.lean|

    \begin{theorem}
        We have for all $x\in G\setminus G'$
        \begin{equation}
            \label{eq-sump} \sum _{{\mathfrak p}\in {\mathfrak P}}T_{{\mathfrak p}} f(x)= \sum _{s=\sigma _1(x)}^{\sigma _2(x)} \int K_{s}(x,y) f(y) e({Q}(x)(y)-{Q}(x)(x))\,  d\mu (y).
        \end{equation}
    \end{theorem}

    \textbf{Formal statement:}
    \begin{lstlisting}
theorem tile_sum_operator {G' : Set X} {f : X → ℂ}
  {x : X} (hx : x ∈ G \ G') : ∑ (p : P X), carlesonOn p f x =
  ∑ s in Icc (σ₁ x) (σ₂ x), ∫ y, Ks s x y * f y * exp (I * (Q x y - Q x x))
\end{lstlisting}

    \textbf{Informal proof:}

    Fix $x\in G\setminus G'$.
    Sorting the tiles ${\mathfrak p}$ on the left-hand-side of \eqref{eq-sump} by the value ${\mathrm{s}}({\mathfrak p})\in [-S,S]$, it suffices to
    prove for every $-S\le s\le S$ that
    \begin{equation}
        \label{outsump} \sum {{\mathfrak p}\in {\mathfrak P}: {\mathrm{s}}({\mathfrak p})=s}T{{\mathfrak p}} f(x)=0
    \end{equation}
    if $s\not\in [\sigma _1(x), \sigma 2(x)]$ and
    \begin{equation}
        \label{insump} \sum {{\mathfrak p}\in {\mathfrak P}: {\mathrm{s}}({\mathfrak p})=s}T{{\mathfrak p}} f(x)= \int K{s}(x,y) f(y) e({Q}(x)(y) - {Q}(x)(x)),  d\mu (y).
    \end{equation}
    if $s\in [\sigma _1(x),\sigma _2(x)]$. If $s\not\in [\sigma _1(x), \sigma 2(x)]$, then by
    definition of $E({\mathfrak p})$ we have $x\not\in E({\mathfrak p})$ for any ${\mathfrak p}$ with ${\mathrm{s}}({\mathfrak p})=s$ and thus $T{{\mathfrak p}} f(x)=0$. This
    proves \eqref{outsump}. Now assume $s\in [\sigma _1(x),\sigma 2(x)]$. By \verb|coverdyadic|, \verb|subsetmaxcube|, \verb|eq-vol-sp-cube|, the fact that $c(I_0) = o$ and
    $G\subset B(o,\frac14 D^S)$, there is at least one $I\in \mathcal{D}$ with $s(I)=s$ and $x\in I$. By \verb|dyadicproperty|, this $I$ is unique. By \verb|eq-dis-freq-cover|,
    there is precisely one ${\mathfrak p}\in {\mathfrak P}(I)$ such that ${Q}(x)\in {\Omega }({\mathfrak p})$. Hence there is precisely one ${\mathfrak p}\in {\mathfrak P}$ with
    ${\mathrm{s}}({\mathfrak p})=s$ such that $x\in E({\mathfrak p})$. For this ${\mathfrak p}$, the value $T{{\mathfrak p}}(x)$ by its definition in \verb|definetp| equals the
    right-hand side of \eqref{insump}. This proves the lemma.

    \textbf{Formal proof:}
    \begin{lstlisting}
rw [P_biUnion, Finset.sum_biUnion]; swap
· exact fun s _ s' _ hss' A hAs hAs' p pA ↦ False.elim <| hss' (s_eq (hAs pA) ▸ s_eq (hAs' pA))
rw [← (Icc (-S : ℤ) S).toFinset.sum_filter_add_sum_filter_not (fun s ↦ s ∈ Icc (σ₁ x) (σ₂ x))]
rw [Finset.sum_eq_zero sum_eq_zero_of_nmem_Icc, add_zero]
refine Finset.sum_congr (Finset.ext fun s ↦ ⟨fun hs ↦ ?_, fun hs ↦ ?_⟩) (fun s hs ↦ ?_)
· rw [Finset.mem_filter, ← mem_toFinset] at hs
exact hs.2
· rw [mem_toFinset] at hs
rw [toFinset_Icc, Finset.mem_filter]
exact ⟨Finset.mem_Icc.2 (Icc_σ_subset_Icc_S hs), hs⟩
· rcases exists_Grid hx.1 hs with ⟨I, Is, xI⟩
obtain ⟨p, IpI, Qp⟩ : ∃ (p : P X), I p = I ∧ Q x ∈ Ω p := by simpa using biUnion_Ω ⟨x, rfl⟩
have pPXs : p ∈ PX_s s := by simpa [s, IpI]
have : ∀ p' ∈ PX_s s, p' ≠ p → carlesonOn p' f x = 0 := by
  intro p' p'PXs p'p
  apply indicator_of_not_mem
  simp only [E, mem_setOf_eq, not_and]
  refine fun x_in_Pp' Qp' ↦ False.elim ?_
  have s_eq := s_eq pPXs ▸ s_eq p'PXs
  have : ¬ Disjoint (I p' : Set X) (I p : Set X) := not_disjoint_iff.2 ⟨x, x_in_Ip', IpI ▸ xI⟩
  exact disjoint_left.1 (disjoint_Ω p'p <| Or.resolve_right (eq_or_disjoint s_eq) this) Qp' Qp
rw [Finset.sum_eq_single_of_mem p pPXs this]
have xEp : x ∈ E p :=
  ⟨IpI ▸ xI, Qp, by simpa only [toFinset_Icc, Finset.mem_Icc, s_eq pPXs] using hs⟩
simp_rw [carlesonOn_def', indicator_of_mem xEp, s_eq pPXs]
\end{lstlisting}
\end{tcolorbox}

\begin{tcolorbox}[title=PFR sample - Example of a relatively uninformative informal proof without broader context, float*,width=\textwidth]

    \textbf{Name:} \verb|cond-trial-ent|

    \textbf{File:} \verb|PFR/ForMathlib/Entropy/Basic.lean|

    \begin{theorem}
        If `X, Y` are conditionally independent over `Z`, then `H[X, Y, Z] = H[X, Z] + H[Y, Z] - H[Z]`.
    \end{theorem}

    \textbf{Formal statement:}
    \begin{lstlisting}
lemma ent_of_cond_indep (hX : Measurable X) (hY : Measurable Y) (hZ : Measurable Z)
     (h : CondIndepFun X Y Z μ) [IsZeroOrProbabilityMeasure μ]
     [FiniteRange X] [FiniteRange Y] [FiniteRange Z] :
     H[⟨X, ⟨Y, Z⟩⟩ ; μ] = H[⟨X, Z⟩; μ] + H[⟨Y, Z⟩; μ] - H[Z; μ]
\end{lstlisting}

    \textbf{Informal proof:}

    Immediate from \verb|conditional-vanish| and \verb|conditional-mutual-alt|.

    \textbf{Formal proof:}
    \begin{lstlisting}
have hI : I[X : Y | Z ; μ] = 0 := (condMutualInfo_eq_zero hX hY).mpr h
rw [condMutualInfo_eq hX hY hZ] at hI
rw [entropy_assoc hX hY hZ, chain_rule _ (hX.prod_mk hY) hZ, chain_rule _ hX hZ, chain_rule _ hY hZ]
linarith [hI]
\end{lstlisting}
\end{tcolorbox}
\twocolumn

\end{document}